\documentclass[final]{cvpr}

\usepackage{times}
\usepackage{epsfig}
\usepackage{graphicx}
\usepackage{cite}
\usepackage{amsmath,amssymb,amsfonts}
\usepackage{algorithmic}
\usepackage{algorithm}
\usepackage{textcomp}
\usepackage{subcaption}
\usepackage{textcomp}
\usepackage{xcolor}
\usepackage{mathtools}
\usepackage{tabularx}
\usepackage{makecell}
\usepackage{dsfont}
\usepackage{balance}
\usepackage{textpos}

\newcommand\blfootnote[1]{%
  \begingroup
  \renewcommand\thefootnote{}\footnote{#1}%
  \addtocounter{footnote}{-1}%
  \endgroup
}

\usepackage[pagebackref=true,breaklinks=true,colorlinks,bookmarks=false,     linkcolor=red, urlcolor=orange,
, citecolor=green]{hyperref}

\def\cvprPaperID{21} 
\def\confYear{CVPR 2021}

\begin{document}

\title{Altruistic Maneuver Planning for Cooperative Autonomous Vehicles Using Multi-agent Advantage Actor-Critic}


\author{Behrad Toghi$^{1,2}$, Rodolfo Valiente$^{1}$, Dorsa Sadigh$^2$, Ramtin Pedarsani$^3$, Yaser P. Fallah$^1$\\
$^{1}$ Connected \& Autonomous Vehicle Research Lab (CAVREL), University of Central Florida, Orlando, FL\\
$^{2}$ Intelligent and Interactive Autonomous Systems Group (ILIAD), Stanford University, Stanford, CA\\
$^{3}$ Department of Electrical and Computer Engineering, University of California, Santa Barbara, CA\\
\texttt{toghi@knights.ucf.edu}}

\maketitle

\begin{abstract}
With the adoption of autonomous vehicles on our roads, we will witness a mixed-autonomy environment where autonomous and human-driven vehicles must learn to co-exist by sharing the same road infrastructure. To attain socially-desirable behaviors, autonomous vehicles must be instructed to consider the utility of other vehicles around them in their decision-making process. Particularly, we study the maneuver planning problem for autonomous vehicles and investigate how a decentralized reward structure can induce altruism in their behavior and incentivize them to account for the interest of other autonomous and human-driven vehicles. This is a challenging problem due to the ambiguity of a human driver's willingness to cooperate with an autonomous vehicle. Thus, in contrast with the existing works which rely on behavior models of human drivers, we take an end-to-end approach and let the autonomous agents to implicitly learn the decision-making process of human drivers only from experience. We introduce a multi-agent variant of the synchronous Advantage Actor-Critic (A2C) algorithm and train agents that coordinate with each other and can affect the behavior of human drivers to improve traffic flow and safety.
\end{abstract}

\begin{textblock*}{15cm}(1.2cm,-19.4cm) 
    \centering
  \textit{Accepted to 2021 IEEE/CVF Conference on Computer Vision and Pattern Recognition (CVPR 2021)\\ \href{https://www.adp3.org/}{Workshop on Autonomous Driving: Perception, Prediction and Planning}}
\end{textblock*}
%
\section{Introduction}
\label{sec:intro}

\blfootnote{Authors B. Toghi and R. Valiente contributed equally.}\blfootnote{This material is based upon work supported by the National Science Foundation under Grant No. CNS-1932037.}Autonomous vehicles (AVs) can leverage their superior computation power, precision, and reaction time to avoid errors occurred by human drivers and drive more efficiently. Connecting AVs and human-driven vehicles (HVs) via vehicle-to-vehicle (V2V) communication creates an opportunity for extended situational awareness and enhanced decision making~\cite{toghi2019analysis, toghi2019spatio}. We are particularly interested in the problem of cooperative decision-making in mixed-autonomy environments where AVs need to share the road infrastructure with human drivers. In such environments, a given AV interacts with other vehicles, whether autonomous or human-driven, and most likely faces conflictive and competitive scenarios where its individual interest does not necessarily align with that of other vehicles.

\begin{figure*}[t]
  \centering
  \includegraphics[width=.96\textwidth]{   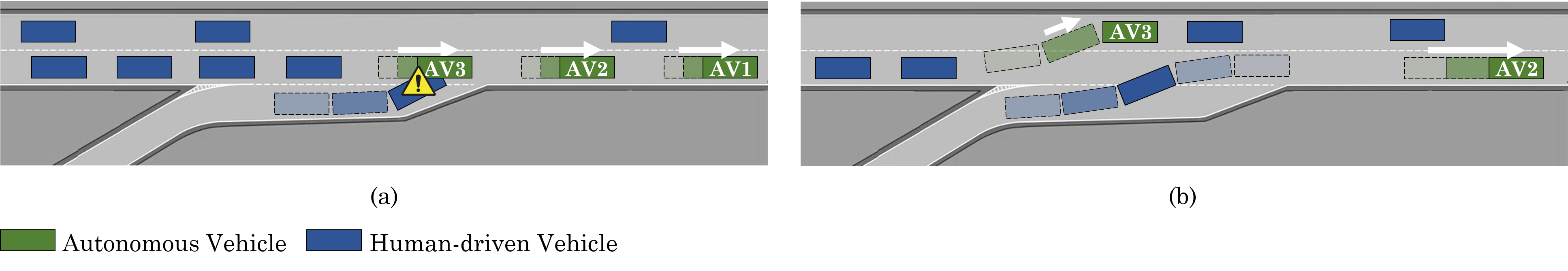}
  \caption{\small{Altruistic AVs can compromise on their individual utility to result socially-desirable behaviors that account for other vehicles. \textbf{\emph{(a)}} Egoistic AVs solely optimize for their own utility and do not allow the merging vehicle to merge, \textbf{\emph{(b)}} Altruistic AVs compromise on their individual utility in order to account for the human-driven vehicles and open up space for the merging vehicle. 
}}
  \label{fig:coopvsnoncoop}
\end{figure*}
%
In his classic, \emph{"I, Robot"}, Isaac Asimov points out these conflicts and lays down the Three Laws of Robotics which establish a non-formal scheme for the co-existence of robots and humans. These rules prescribe principles that an intelligent agent needs to follow in order to protect its own existence and utility while keeping the involving humans safe. The first law emphasizes the importance of protecting humans, \emph{"a robot may not injure a human being or, through inaction, allow a human being to come to harm"}, and the third law defines the local interest of a robot, \emph{"A robot must protect its own existence as long as such protection does not conflict with the previous laws"}~\cite{asimov2004robot}. Despite being rooted in science fiction, this set of rules later inspired the debates on the ethics of artificial intelligence and robotics~\cite{lin2011robot} and is meaningfully relevant in our particular problem. Conflicts and competitions naturally arise in a mixed-autonomy driving environment and it is crucial to assemble cooperative decision-making criteria that ensure safety and efficiency for both human-driven and autonomous vehicles.

The example in Figure~\ref{fig:coopvsnoncoop} helps us to elaborate on our mixed-autonomy driving problem: a vehicle that is attempting to merge into a highway requires the cruising vehicles to open up space for it. Here, the individual interest of the cruising vehicles does not align with that of the merging vehicle and therefore egoistic decision-making will not necessarily lead to an outcome that is optimal for the group of cars. With altruistic decision making, vehicles can optimize for a combination of their individual utility and a social utility. Our key insight is that by working together, a group of vehicles can overcome the physical limitations of a single vehicle and achieve socially-desirable outcomes~\cite{palanisamy2020multi, sadigh2016planning}. However, we believe the willingness of human drivers in demonstrating selflessness and altruism is rather ambiguous and thus cannot be taken for granted. Instead, we rely on the autonomous agents to shape and guide the traffic flow in a fashion that optimizes the social objective.

In order to achieve the above goal, an intelligent agent needs to 1) be able to understand and predict the behavior of human drivers and 2) coordinate with its allies, i.e., other AVs, to construct formations that eventually benefit the group. Understanding and anticipating the actions of human drivers is a challenging task as it is often difficult to model human behaviors. The models that keep track of beliefs are mostly not scalable and partner modeling techniques predominantly are intractable in capturing a belief over partners' behaviors that often vary due to factors such as fatigue, distraction, etc.~\cite{xie2020learning}. The core differentiating idea that we emphasize on is that the human driver models which are extracted in the absence of autonomous agents, are not necessarily valid when humans confront autonomous agents. To address these challenges, we take an end-to-end approach and induce altruism into the decision-making process of autonomous agents using a decentralized reinforcement structure. Despite not having access to an explicit model of the human drivers, the trained autonomous agents learn to implicitly model the environment dynamics, including the behavior of human drivers, which enables them to interact with HVs and guide their behavior. 

Through our experiments in a highway merging scenario, we demonstrate that our agents are able to learn from scratch not only to drive, but also understand the behavior of HVs and coordinate with them. Our main contributions are as follows:
\begin{itemize}
  \item Using a multi-agent variant of Advantage Actor-Critic (A2C), we implement a decentralized reinforcement learning algorithm that induces altruism into the decision-making process of autonomous vehicles and incentivizes them to account for the utility of other vehicles in order to result socially-desirable behaviors.
  \item We choose a complex and conflictive driving scenario and show that our altruistic autonomous agents are able to plan long sequences of actions that reduce the number of crashes and improve the overall traffic flow as compared to egoistic autonomous agents.
\end{itemize}

\section{Literature Review}
\label{sec:related_work}

A fundamental challenge in multi-agent learning and training agents that evolve concurrently is the intrinsic non-stationarity of the environment. Some approaches to address this issue, such as the work by Arel~\etal, assume that all agents have access to the global state of the world or that they could share their states among the neighbors~\cite{arel2010reinforcement}. However, these assumptions are not always practical in real-world problems. Egorov~\etal attempt to overcome this challenge using a centralized Critic function that can mitigate the effect of non-stationarity in the learning process~\cite{egorov2016multi}. Foerster~\etal propose the counterfactual multi-agent (COMA) algorithm which utilizes a set of joint actions of all agents as well as the full state of the world during the training~\cite{foerster2018counterfactual}. A global centralized reward function is then used to calculate the agent-specific \textit{advantage function}. In contrast, we assume partial observability and a decentralized reward function during both training and execution that is expected to promote cooperative and sympathetic behavior among autonomous vehicles. Lowe~\etal present a general-purpose multi-agent learning algorithm that enables agents to conquer simple cooperative-competitive games with access to local observations of the agents~\cite{lowe2017multi}. An adaptation of Actor-Critic methods with a \textit{centralized action-value function} is employed that uses the set of actions of all agents and local observations as its input. In our work, however, agents do not have access to the actions of their allies and/or opponents.

Within the social navigation literature, Omidshafiei~\etal and Lauer~\etal study the multi-agent navigation and focus on solving cooperative and competitive problems by making assumptions on the nature of interactions between agents (or agents and humans)~\cite{omidshafiei2017deep, lauer2000algorithm}. However, in our case, we are interested in the emerging sympathetic cooperative behavior that enables the agents to cooperate among themselves as well as with their competitors, i.e., humans. Recent works by Pokle~\etal have revealed the potential for collaborative planning and interaction with humans\cite{pokle2019deep}. Alahi~\etal have proposed the social-LSTM framework to learn general human movement and predict their future trajectories~\cite{alahi2016social}. Toghi~\etal present a maneuver-based dataset and model on human driving that can be used to classify driving maneuvers~\cite{toghi2020maneuver}. Nikolaidis~\etal optimize a common reward function in order to enable joint trajectory planning for humans and robots~\cite{nikolaidis2015efficient}. In contrast, we seek altruistic behaviors without having an explicit model of the human driver's behavior or relying on their aid. Sadigh~\etal also demonstrate an approach based on imitation learning ~\cite{sadigh2016planning} to learn a reward function for the human drivers and then employ that to affect the behavior of human-driven vehicles. Mahjoub~\etal took a statistical machine learning approach to the problem of driver behavior prediction and were able to predict sub-maneuvers within a short time-horizon~\cite{8690570, 8690965, mahjoub2019v2x}. Works by Wu\etal and Lazar\etal study the human-autonomy problem from a macro-traffic perspective and have demonstrated emerging human behaviors within mixed-autonomy scenarios and the possibility to leverage these patterns to control and stabilize the traffic flow~\cite{wu2017emergent, lazar2019learning}.

\section{Problem Statement}
\label{sec:problem}
\begin{figure}[tb]
  \centering
  \includegraphics[width=.49\textwidth]{   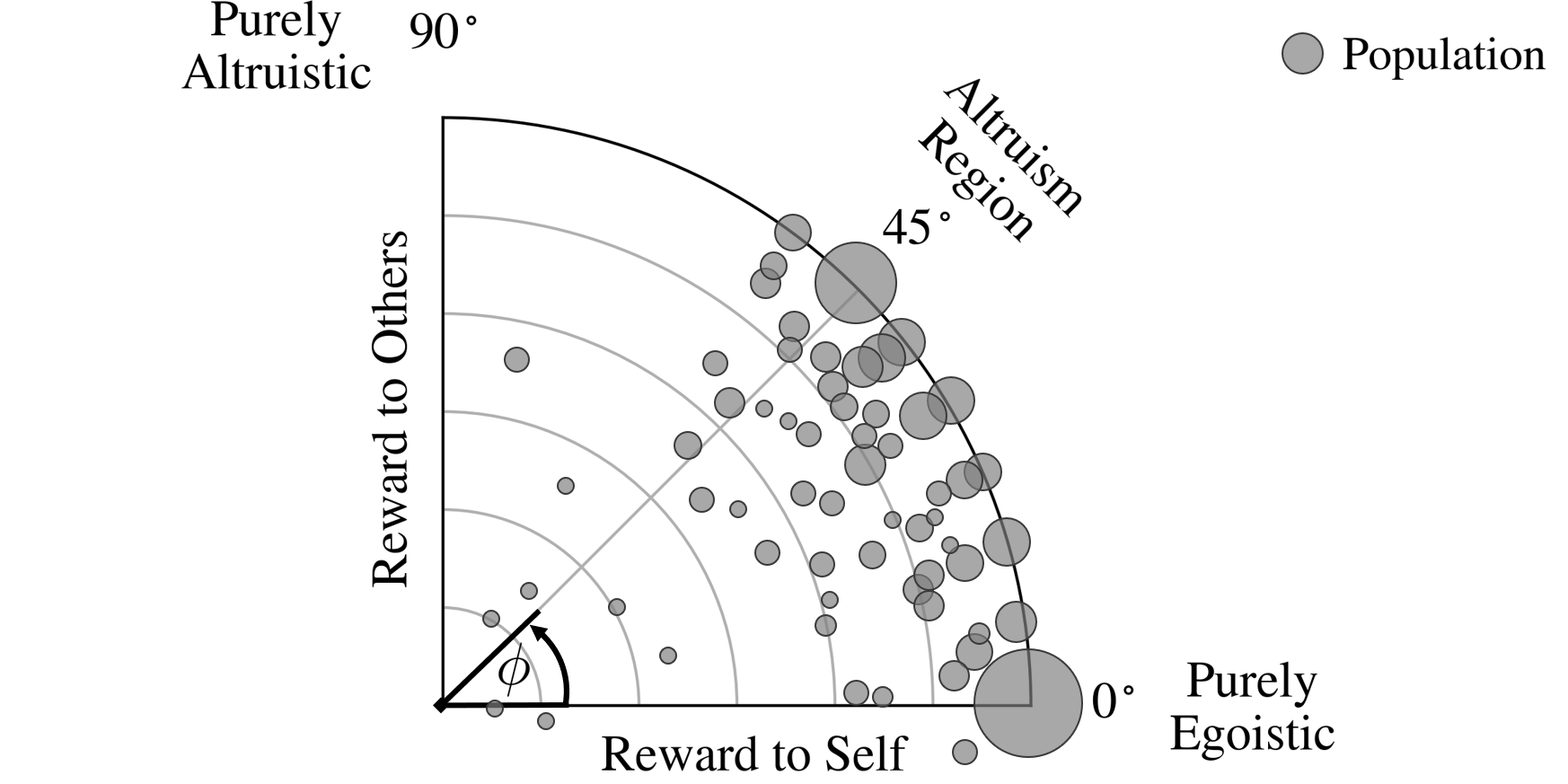}
  \caption{\small{The Social Value Orientation ring demonstrates different behaviors based on a human/robot's preference to account for others. The diameter of the circles show the likelihood of a specific behavior. Figure is based on the data from~\cite{garapin2015does}}}
  \label{fig:svo}
\end{figure}
Humans and intelligent agents interacting in a mixed-autonomy environment can hold different levels of altruism. The preference of a person to account for the interest of others in spending its individual resources, e.g. time, money, energy, is formally studied as that person's Social Value Orientation (SVO) in social psychology~\cite{schwarting2019social, garapin2015does}. This concept can also be employed to quantify an autonomous agent's willingness to act egoistically or altruistically~\cite{buckman2019sharing}. As demonstrated in Figure~\ref{fig:svo}, the behavior of a human, or similarly an autonomous agent, can lay on anywhere from absolutely egoistic to absolutely altruistic depending on the weight they assign to the utility of others. In order to achieve a set of socially-desirable maneuvers, autonomous agents must act based on the decisions made by their allies, i.e., other AVs, and human drivers.

It is well established in the behavioral decision theory that the SVO of humans is rather heterogeneous. Murphy and Ackermann~\cite{murphy2015social} formalize the human decision-making process as maximizing an expected utility with weights governed by the human's individual preferences. However, these preferences are typically unknown which makes the SVO of a human ambiguous and unclear. The existing works on social navigation for AVs and robots in general, often make assumptions on the humans' willingness to cooperate with autonomous agents, whereas Figure~\ref{fig:svo} points out the large likelihood of having an egoistic human with no will to account for others. Therefore, we define our problem in a worst-case scenario configuration and assume that all human drivers are egoistic and hence cannot be relied on in terms of cooperating with autonomous agents. Autonomous agents must coordinate with each other and when it is necessary, shape the behavior of human drivers around them to realize a social goal that benefits the group of the vehicle. The desired outcome for our particular driving problem would be enabling seamless and safe highway merging while maximizing the distance traveled by all vehicles and obviously, avoiding collisions.

\begin{figure}[tb]
  \centering
  \includegraphics[width=.44\textwidth]{   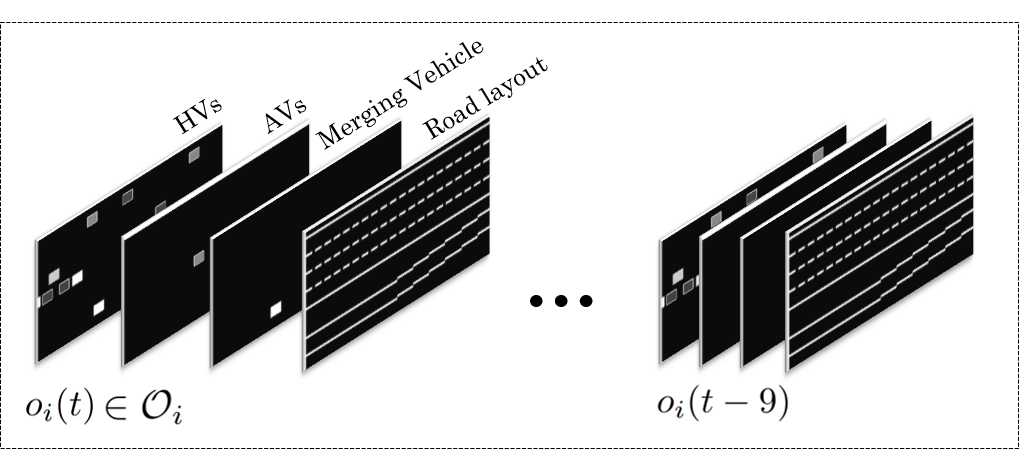}
  \caption{\small{Stacked multi-channel VelocityMap state representation embeds the speed and position of vehicles. Each observation $o_i$ is a tensor of size $10 \times (4 \times 64 \times 512$).}}
  \label{fig:velocitymap}
\end{figure}

\section{Proposed Solution}
\label{sec:solution}
Our key insight is bypassing the need for prior information on human drivers' behavior and relying on them to cooperate with AVs. Instead, we adjust the AVs' SVO through a decentralized reinforcement learning algorithm which promotes altruistic behavior in their decision-making process. As opposed to the existing works which rely on derived behavior models of human drivers, we follow our methodology in~\cite{toghi2021cooperative, toghi2021social} which allows the autonomous agents to implicitly model human drivers SVO in real-time and through experience.

\subsection{Formalism}
\label{sec:formalism}
Consider a highway merging ramp and a straight highway with a group of human-driven and autonomous vehicles cruising on it, similar to what is depicted in Figure~\ref{fig:coopvsnoncoop}. All vehicles are connected together using V2V communication are able to share their observations~\cite{toghi2018multiple, shah2019real, saifuddin2020performance}. The highway merging scenario is particularly interesting to us due to its competitive nature as the interest of the merging vehicle does not align with the interest of the cruising vehicles. We formally annotate a set of autonomous vehicles $\mathcal{I}$, a set of human-driven vehicles $\mathcal{V}$, and the human-driven merging vehicle, $M \in \mathcal{V}$ which is attempting to merge into the highway. AVs share their situational awareness through cooperative perception to overcome the limitations of occluded and non-line-of-sight vision~\cite{valiente2019book, valiente2019controlling}. Thus, we assume each AV has a local partial observation of the world constructed using the shared situational awareness with its allies and is able to observe a subset of AVs $\widetilde{\mathcal{I}} \subset \mathcal{I}$, in addition to a subset of HVs $\widetilde{\mathcal{V}} \subset \mathcal{V}$.
\begin{figure}[tb]
  \centering
  \includegraphics[width=.48\textwidth]{   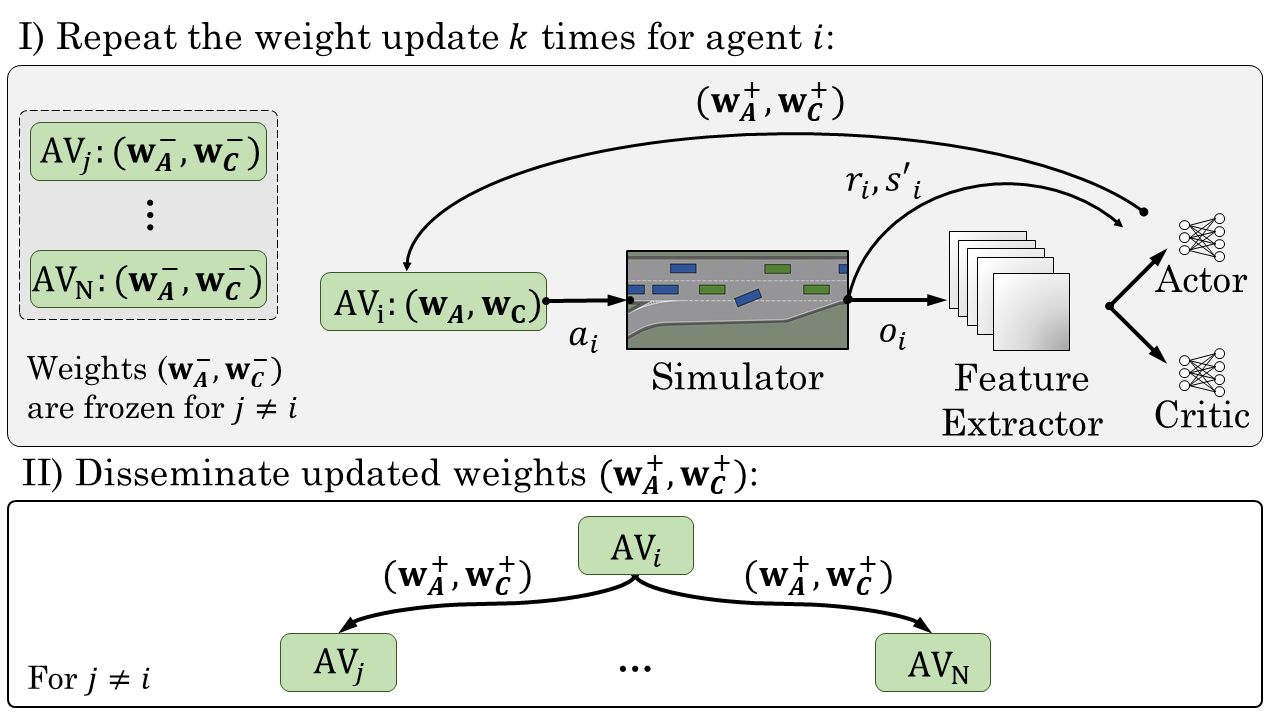}
  \caption{\small{The multi-agent Advantage Actor-Critic framework and policy dissemination process.}}
  \label{fig:marl_policy}
\end{figure}
\subsection{Observations and Actions}
As mentioned before, our interest is in maneuver planning for AVs and studying their interaction with human drivers. This fact motivates us to view the problem from an abstracted view that determines the meta-actions for an autonomous agent. These meta-actions then further get translated into optimized trajectories and then low-level control signals, e.g., throttle level and steering angle. Specifically, we define an AV's action-space as a set of discrete meta-actions $a_i$:
\begin{equation}
\label{equ:action_space}
    a_i \in \mathcal{A}_i =
    \begin{bmatrix}
        \texttt{Lane Left}\\
        \texttt{Idle}\\
        \texttt{Lane Right}\\
        \texttt{Accelerate}\\
        \texttt{Decelerate}
    \end{bmatrix}
\end{equation}

We design a \textit{multi-channel VelocityMap} embedding, as shown in Figure~\ref{fig:velocitymap}, to represent the state of an autonomous agent. VelocityMaps embed the relative position of AVs and HVs into two channels with pixel values proportional to their relative speed. Additionally, we embed the road layout, position of the observer, and the relative position of the merging vehicle (only if it fits within the observer's perception range) in separate layers. Driving data carries important temporal patterns which can help in modeling human behavior and predictive maneuver planning. Hence, we stack a group of VelocityMaps recorded at consecutive time-steps to capture this temporal information, as Figure~\ref{fig:velocitymap} depicts.

\subsection{Decentralized Reinforcement Structure}
\label{sec:decentralized_reinforcement}
Inspired by the work done on social autonomous driving by Schwarting \etal~\cite{schwarting2019social} we define the decentralized reward function of an autonomous agent, $I_i \in \mathcal{I}$, based on its SVO. For the scenario defined in Section~\ref{sec:formalism}, 
\begin{equation}
\label{equ:reward1}
    \begin{aligned}
    R_i(s_i, a_i; \phi_i) ={} & \cos(\phi_i) R^E_i + \sin(\phi_i) R^A_i
    \\={} & \cos(\phi_i)  r^E_i(s_i, a_i) + \\
    & \sin(\phi_i) \Big[ \sum_j r_{i, j} (s_i, a_i) + \sum_k r_{i, k} (s_i, a_i) \Big]
    \end{aligned}
\end{equation}
where $\phi_i$ is the SVO angular preference of the agent $I_i$ and determines its level of altruism. Summations are over $j \in \widetilde{\mathcal{I}} \setminus \{I_i\}$, $k \in \widetilde{\mathcal{V}} \cup \{M\}$. The individual performance of a vehicle, $r$, can be quantified using metrics such as distance traveled, average speed, and a negative cost for lane-change or acceleration/deceleration to minimize the jerk and unnecessary maneuvers. The altruistic part of the reward, $R^A$, sums over the individual performance of the other vehicles. Using such a decentralized reward function in our multi-agent reinforcement learning setup, we can train agents that optimize for a social utility governed by their SVO, i.e., we induce altruism in their decision-making process without knowledge on the SVO of the humans.

\subsection{Multi-agent Advantage Actor-Critic}
\label{sec:a2c}

\begin{figure}[t]
    \centering
    \includegraphics[width=0.47\textwidth]{   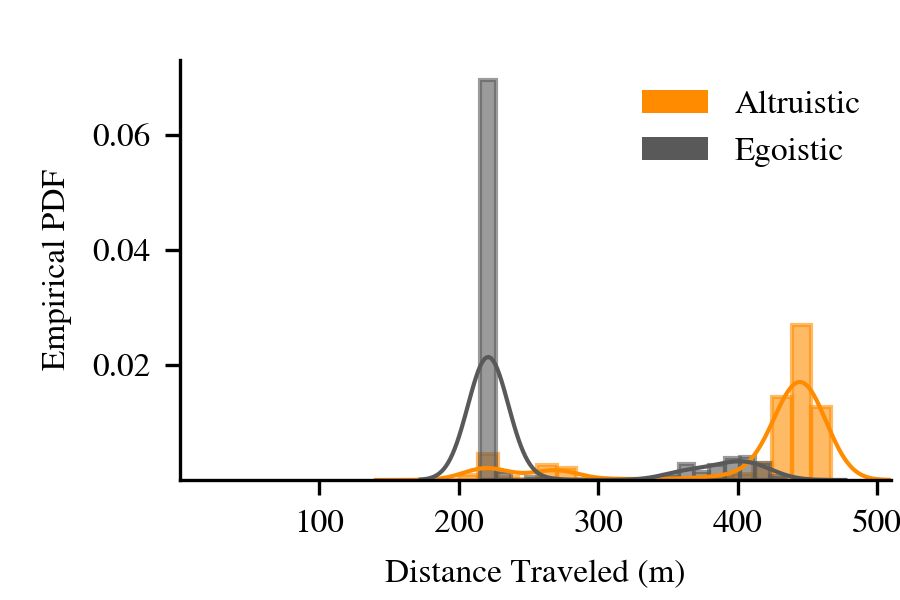}
    \caption{\small{Distribution of distance traveled by the merging vehicle when the AVs act egoistically \emph{(Gray)} as compared to the case with altruistic AVs \emph{(Orange)}.}}
    \label{fig:distance_distribution}
\end{figure}
%

\begin{table}[b]
\caption\small{{Architecture of A2C and Feature Extractor Networks.}}
\begin{center}
 \begin{tabular*}{0.49\textwidth}{c c c c}
  \small{Layer} & \small{Type} & \small{Size} & \small{Activation} \\ 
 \hline\hline
 \multicolumn{4}{l}{\textbf{\small{\textit{Feature Extractor Network}}}} \\ 
 \hline
 \small{0} & \small{Input} &\small{ $10 \times  (4\times 64\times 512)$} & \small{-} \\ 
\small{1} & \small{Conv3D} & \small{$32\,\, @\,\, (1\times 8\times 8)$ } &\small{ReLU} \\
\small{2} & \small{Conv3D} & \small{$64\,\, @\,\, (3\times 4\times 4)$}  &\small{ReLU }\\
\small{3} & \small{Conv3D} &\small{ $64\,\, @\,\, (3\times 3\times 3)$ } & \small{ReLU }\\
\small{4} & \small{FC} & \small{512}  & \small{ReLU }\\
\hline
 \multicolumn{4}{l}{\textbf{\small{\textit{Actor and Critic Networks}}}} \\ 
 \hline
 \small{1} & \small{FC }&\small{ 256 }  & \small{ReLU} \\
 \small{2} & \small{FC} &\small{ 5 }  & \small{Softmax} \\
 \hline
 \hline
 \end{tabular*}
\end{center}
\label{table:Arch}
\end{table}

As we discussed in Section~\ref{sec:problem}, a major challenge in multi-agent reinforcement learning is dealing with the set of learning agents that are evolving concurrently and therefore making the environment non-stationary. Particularly, this creates an issue for the methods based on the experience replay mechanism, such as Deep Q-networks. In order to avoid the need for the replay buffer, we employ a synchronous Advantage Actor-Critic (A2C) algorithm in which the Critic approximates the state-value function, $V(s)$, and the Actor update the stochastic policy based on the gradient from the Critic. The difference between the action-value function, $Q(s,a)$, and $V(s)$ can be used as a baseline, commonly referred to as the Advantage Function~\cite{sutton2018reinforcement}, i.e.,
\begin{equation}
\label{equ:advantagefunction}
    A(s_t, a_t) = r_{t+1} + \gamma V_{\textbf{\text{w}}_C}(s_{t+1})-V_{\textbf{\text{w}}_C}(s_t)
\end{equation}
where $\textbf{\text{w}}_C$ denotes the weights of the neural network for the Critic. Thus, the update rule can be simplified as,
\begin{equation}
\label{equ:a2cupdate}
\nabla_{\theta}J(\theta) \sim  \sum_{t=0}^{T-1} \nabla_\theta \log \pi_\theta (a_t|s_t) A(s_t, a_t)
\end{equation}
where $\pi_\theta$ is the policy parameterized by parameters $\theta$.

As demonstrated in Figure~\ref{fig:marl_policy}, we use a 3D convolutional network (Conv3D) as the feature extractor with the stacked VelocityMaps as its input and two separate multi-layer perceptron networks for the Actor and the Critic. A 3D convolutional architecture enables us to extract the temporal information embedded in the VelocityMaps, details of the feature extractor network are listed in Table~\ref{table:Arch}. The Actor network outputs a probability distribution corresponding to each action. The Critic network maps each state to its corresponding state-value. To stabilize the multi-agent training process, we train each AV in a semi-sequential fashion by freezing the networks of all other AVs $(\textbf{w}_A^-, \textbf{w}_C^-)$, and then disseminating the updated weights, $(\textbf{w}_A^+, \textbf{w}_C^+)$, periodically.

\section{Experiments}
\label{sec:experiments}
We use an abstract traffic simulator for our experiments. An OpenAI Gym environment provided by Leurent~\etal~\cite{leurent2019approximate} is customized to suit our particular driving scenario and multi-agent problem. A Kinematic Bicycle Model governs the kinematics of the vehicles and a proportional–integral–derivative (PID) controller renders the meta-actions into low-level steering and throttle control signals. The dynamics of the HVs and their decision-making process is implemented using the lateral and longitudinal driver models proposed by Treiber~\etal and Kesting~\etal~\cite{treiber2000congested, kesting2007general}.

\begin{figure}[t]
  \centering
  \includegraphics[width=.45\textwidth]{   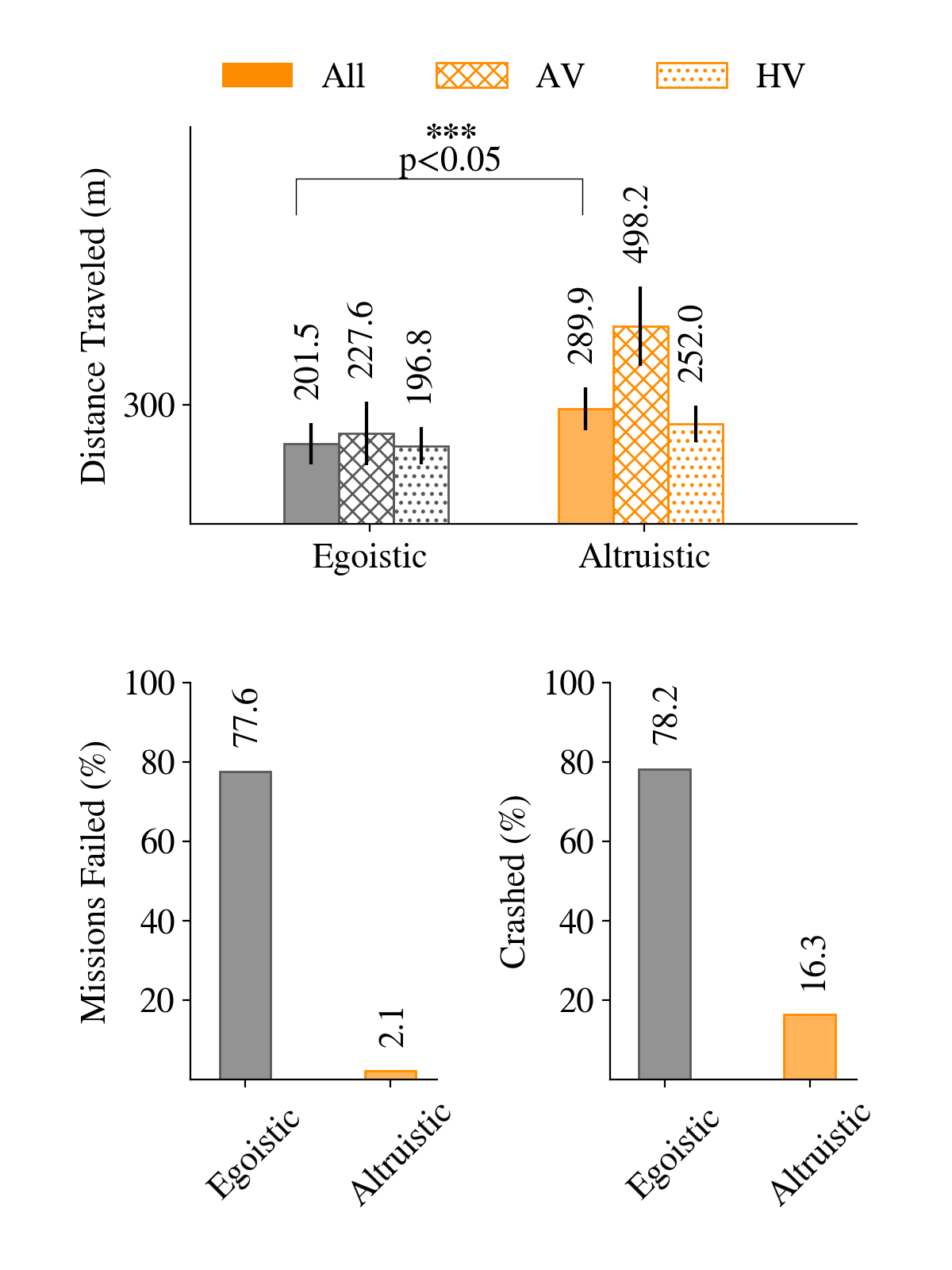}
  \caption{\small{Our experiments demonstrate that our altruistic autonomous agents are able to take sequences of actions that reduce the number of crashes and improve the overall traffic flow as compared to egoistic autonomous agents.}}
  \label{fig:barplots}
\end{figure}
\begin{figure}[b]
    \centering
    \includegraphics[width=0.48\textwidth]{   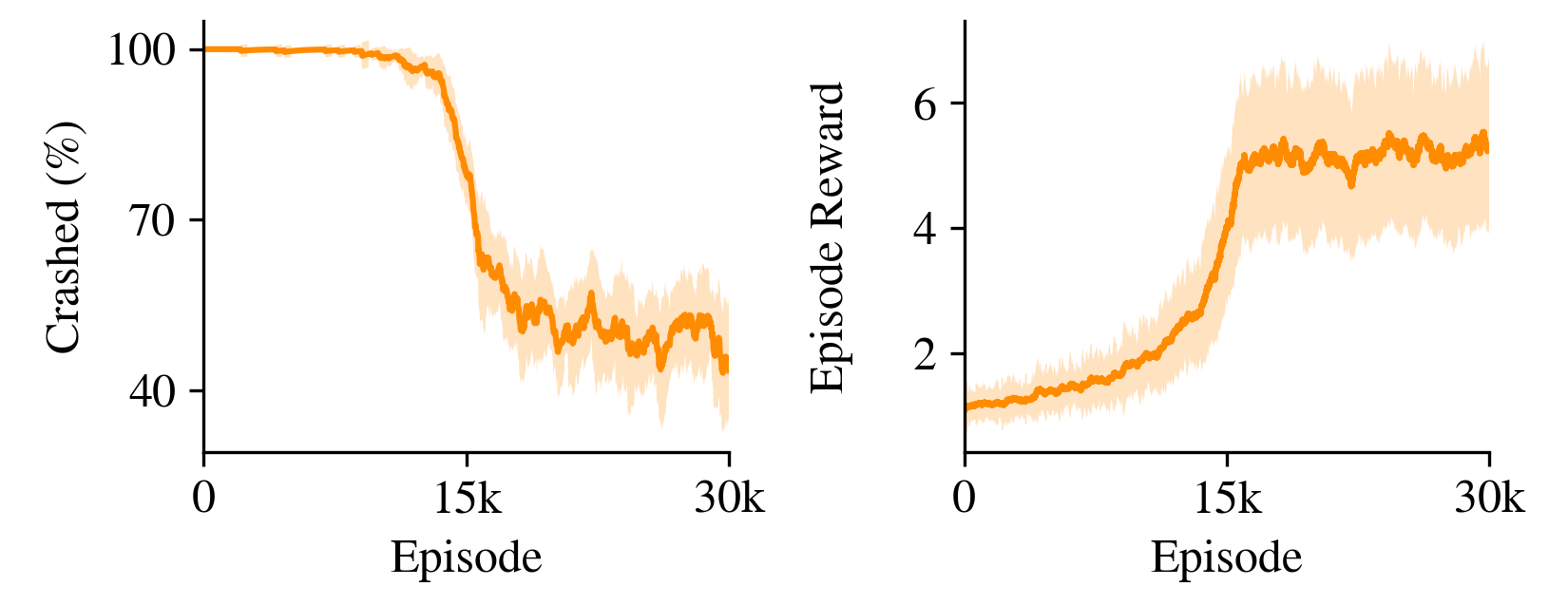}
    \caption{\small{Evolution of our altruistic agents during training using a decentralized reinforcement learning algorithm.}}
    \label{fig:trainingsummary}
\end{figure}
\noindent \textbf{Manipulated Factors. }
Equation~\ref{equ:reward1} characterizes the decentralized reward that an AV, $I_i \in \mathcal{I}$ receives by taking an action $a_i$ at the state $s_i$. The agent's SVO angular preference, $\phi_i$, determines the importance that AV is willing to put on other vehicles' utility. Thus, we experiment with the values of $\phi_i=0$, i.e., purely egoistic behavior, and $\phi_i=\pi/4$, i.e., altruistic behavior, to examine the benefits of altruistic maneuver planning in comparison with AVs that act egoistically.

\noindent \textbf{Performance Metrics. }
As discussed in Section~\ref{sec:intro}, we aim to leverage the altruistic behavior of AVs to 1) enhance the overall traffic flow, 2) improve driving safety, and last but not least 3) generate socially-desirable outcomes. Hence, we gauge our results using an intuitive traffic flow metric that shows the average distance traveled by all vehicles in the experiment. The percentage of episodes with a collision in them is also measured to assess the safety aspect of the resulting behaviors. Episodes with a successful and seamless merging are also counted to provide an understanding of the AVs' ability to consider social utility.

\begin{figure*}[t]
  \centering
  \includegraphics[width=.9\textwidth]{   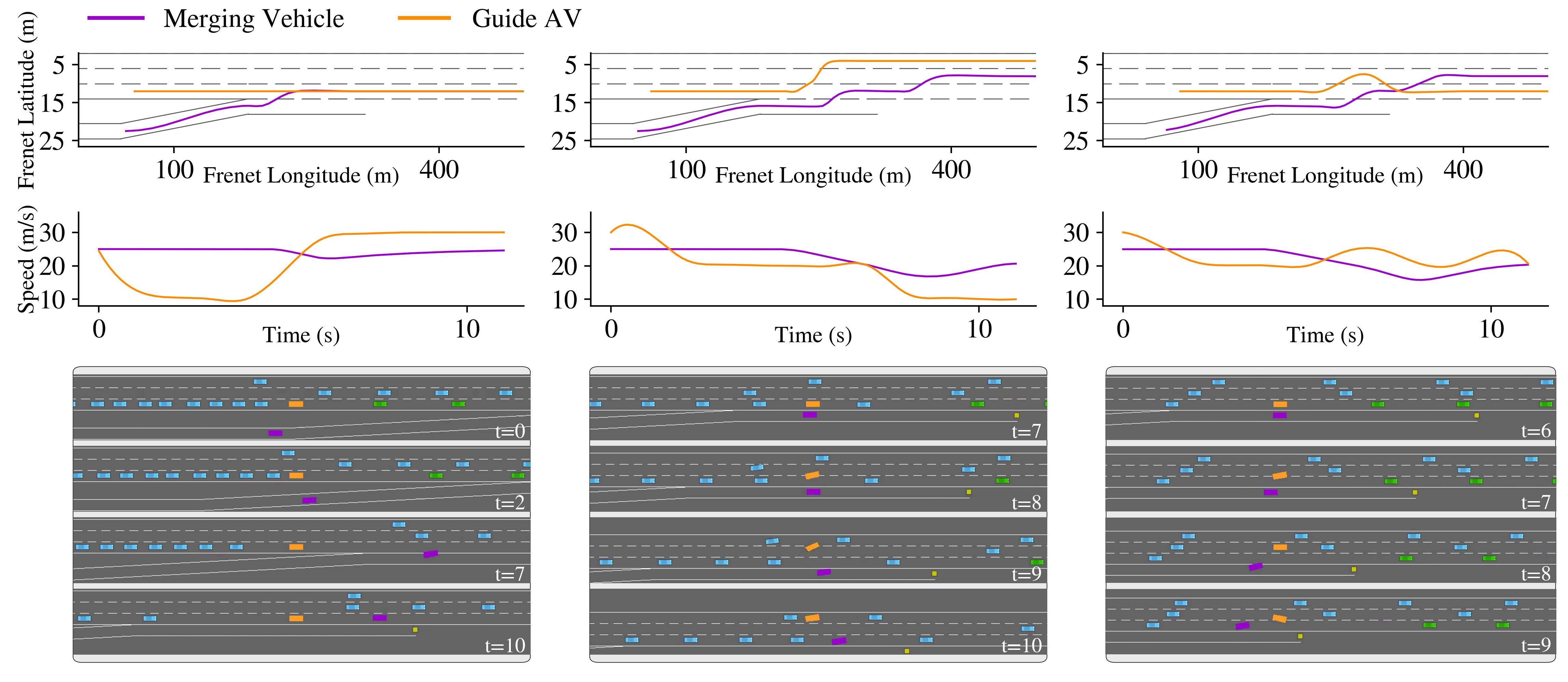}
  \caption{\small{Three example training episodes demonstrate how altruistic AVs learn to take a sequence of decisions that eventually lead to safe, efficient, and socially-desirable outcomes. AVs and HVs are annotated with green and blue color, respectively.}}
  \label{fig:trajectories}
\end{figure*}
\noindent \textbf{Hypothesis. }
We examine a key hypothesis that we initially stated in Section~\ref{sec:intro}. By varying the SVO angular preference towards the purely altruistic behavior, i.e., $\phi=\pi/2$, AVs learn to work together and affect the behavior of human drivers to open up a safe corridor for the merging vehicle and allow it to merge safely and seamlessly. We hypothesize despite some AVs compromising on their individual utility and acting altruistically, the overall traffic-level metrics improve as a result of their altruistic behavior. More importantly, we believe that our multi-agent A2C algorithm enables AVs to take socially desirable sequences of actions.

\noindent \textbf{Findings. }
Using the proposed multi-agent A2C algorithm structure explained in Section~\ref{sec:a2c}, we train the autonomous agents for 30,000 episodes and evaluate their performance over 3,000 test episodes. The learning performance and the evolution of our metrics during the training are shown in Figure~\ref{fig:trainingsummary}. We ran multiple rounds of training to ensure that agents converge to similar points in terms of training stability.

The overall performance of both egoistically and altruistically trained autonomous agents is demonstrated in Figure~\ref{fig:barplots}. Confirming our hypothesis, we can observe that altruistic AVs manage to significantly reduce the number of collisions while improving the average distance traveled by all vehicles, which expresses the improved traffic flow. More importantly, it is clear that egoistic AVs did not allow the merging vehicle to safely merge into the highway and caused sub-optimal and unsafe situations, whereas altruistic AVs worked together in order to open up a safe space for the merging HV and allow it to merge.

To grasp a better understanding of how altruism changes the decision-making process of AVs, we dive deeper into the results and particularly measure the average distance traveled by the merging vehicle in both altruistic and egoistic settings. Figure~\ref{fig:distance_distribution} illustrates the empirical distribution of the distance traveled by the merging vehicle. It is evident that when AVs are egoistic, in the majority of episodes the merging vehicle either crashes after traveling $\sim220m$ or gets stuck in the merging ramp and hence not able to move. In contrast, in the episodes that contain altruistic AVs, the merging vehicle is able to safely merge into the highway and consequently travel approximately as twice as the previous case, $\sim450$, and reach farther distances on the highway.

We narrow down our investigation particularly to the behavior of the merging vehicle and the AV that directly meets the merging vehicle at the merging point, i.e., the \emph{Guide AV}. Although the merging vehicle mostly has a direct interaction only with the vehicles that are cruising on the right-most lane of the highway but the existence of other vehicles and their interactions, indirectly affects the sequence of actions that both the merging vehicle and the Guide AV take. We deliberately perform our study in a multi-lane highway with a fairly large mixed population of AVs and HVs to investigate these indirect relations. It is important to note that although the size of the discrete action-space defined in Equation~\ref{equ:action_space} is relatively small but a vehicle in such a complex multi-lane highway faces a large combination of sequences of actions that it can take.

The eventual outcome of an episode is determined by the sequence of maneuvers taken by vehicles and not their momentarily actions. Thus, an AV must be able to make decisions that lead to a desirable outcome in time-steps ahead. In Figure~\ref{fig:trajectories}, we illustrate three sample behaviors recorded during the training of autonomous agents. In the first example (the left-most plots), the Guide AV (orange) sees a group of HVs (blue) blocking its left-hand-side lane and thus has no choice but to stay in its lane and slow down to block the traffic behind and open up space for the merging vehicle (purple). This behavior is observable in the speed profile of the Guide AV as well as the provided snapshots. In the second example (the middle plots), the Guide AV's path is blocked by another HV but it finds the left lane open and thus makes a lane-change to open up space for the merging vehicle, resulting in a safe merging. In the third example (the right-most plots), the Guide AV has mastered a way to both enable the merging vehicle to merge, i.e., account for the utility of others, and also perform another lane-change to speed up and optimize for its local utility.

\section{Concluding Remarks}
Overall, we observe that using our proposed decentralized multi-agent learning scheme, we are able to induce altruism into the decision-making process of autonomous vehicles and adjust their social value orientation. Our altruistic agents not only learn to drive on the highway environment from scratch but also are able to coordinate with each other and affect the behavior of humans around them to realize socially-desirable outcomes that eventually improve traffic safety and efficiency. Our work on social navigation is limited in the sense that we have not used actual human drivers in our simulations nor realistic human driving data and instead have implemented a simple driving model from the literature. However, we believe that our solution can be extended using more realistic human behaviors and our trained agents should be able to adapt accordingly. Additionally, we used a simple remedy to grasp the temporal dependencies in driving data using our stacked VelocityMaps, however, more sophisticated solutions such as recurrent architectures can further improve the capabilities of our methods in maneuver planning.

{\small
\bibliographystyle{ieee_fullname}
\bibliography{egbib}
}

\end{document}